\documentclass{article} 
\usepackage{iclr2026_conference,times}

\usepackage{amsmath,amsfonts,bm}

\def\eqref#1{equation~\ref{#1}}

\def\1{\bm{1}}

\DeclareMathAlphabet{\mathsfit}{\encodingdefault}{\sfdefault}{m}{sl}
\SetMathAlphabet{\mathsfit}{bold}{\encodingdefault}{\sfdefault}{bx}{n}

\usepackage[utf8]{inputenc} 
\usepackage[T1]{fontenc}    
\usepackage{amsmath}  
\usepackage[table]{xcolor} 
\definecolor{Highlight}{rgb}{0.21,0.49,0.74}
\definecolor{red}{HTML}{cc1100}
\usepackage[pagebackref,breaklinks,colorlinks=true,linkcolor=red,citecolor=Highlight]{hyperref}       
\usepackage{xcolor}         
\usepackage[capitalize]{cleveref} 
\usepackage{graphicx} 
\usepackage{wrapfig} 
\usepackage{float} 
\usepackage{caption} 
\usepackage{multirow} 
\usepackage{url}            
\usepackage{booktabs}       
\usepackage{amsfonts}       
\usepackage{nicefrac}       
\usepackage{microtype}      
\usepackage{array}         
\usepackage{subcaption}
\usepackage{siunitx} 

\usepackage[ruled,vlined]{algorithm2e} 
\usepackage{amsmath}                   
\usepackage{amssymb}                   
\Crefname{algocf}{Algorithm}{Algorithms}

\usepackage{tcolorbox}      

\newlength\savewidth
\newcommand{\tablestyle}[2]{\setlength{\tabcolsep}{#1}\renewcommand{\arraystretch}{#2}\centering\footnotesize}
\renewcommand{\paragraph}[1]{\vspace{1.25mm}\noindent\textbf{#1}}

\newcolumntype{x}[1]{>{\centering\arraybackslash}p{#1pt}}
\newcolumntype{y}[1]{>{\raggedright\arraybackslash}p{#1pt}}
\newcolumntype{z}[1]{>{\raggedleft\arraybackslash}p{#1pt}}

\newcommand{\ours}{VisionReasoner} 

\usepackage{xcolor}
\definecolor{mycolor}{RGB}{255,87,34}

\usepackage{cleveref} 
\usepackage{marvosym}

\title{{\ours}: Unified Reasoning-Integrated Visual Perception via Reinforcement Learning}
\iclrfinalcopy

\author{%
    Yuqi Liu$^{1*}$ \hspace{1pt}
    Tianyuan Qu$^{1*}$ \hspace{1pt}
    Zhisheng Zhong$^{1}$ \hspace{1pt}
    Bohao Peng$^{1}$ \hspace{1pt}
    Shu Liu$^{2\text{\Letter}}$ \hspace{1pt}
    Bei Yu$^{1}$ \hspace{1pt}
    Jiaya Jia$^{2,3}$ \hspace{1pt}
    \\ 
    CUHK$^{1}$ \hspace{6pt} SmartMore$^{2}$ \hspace{6pt}  HKUST$^{3}$ \hspace{6pt}
     \tt\small {$^*$ Equal contribution \hspace{3pt} $^{\text{\Letter}}$ Corresponding author} \\
    {\tt\small \url{https://github.com/JIA-Lab-research/VisionReasoner}}
}
\begin{document}

\maketitle

\begin{abstract}
Large vision-language models exhibit inherent capabilities to handle diverse visual perception tasks. In this paper, we introduce {\ours}, a unified framework capable of reasoning and solving multiple visual perception tasks within a shared model. Specifically, by designing a unified reward mechanism and multi-object cognitive learning strategies, {\ours} enhances its reasoning capabilities to analyze visual inputs, and addresses diverse perception tasks within a unified model. {\ours} generates a structured reasoning process before delivering the desired outputs responding to user queries. Human evaluation reveals the reasoning process of {\ours} is faithful and reliable even without annotated reasoning train data. To rigorously assess unified visual perception capabilities, we evaluate {\ours} on ten diverse tasks spanning three critical domains: detection, segmentation, and counting. Experimental results show that {\ours} achieves superior performance as a unified model, outperforming the baseline Qwen2.5VL by relative margins of 29.1\% on COCO (detection), 22.1\% on ReasonSeg (segmentation), and 13.2\% on CountBench (counting). 
\end{abstract}

\section{Introduction}

Recent advances in large vision-language models (LVLMs) \citep{bai2025qwen25vl, wang2024qwen2vl, google2025gemini, openai2025gpt41} have demonstrated remarkable capabilities in visual conversations.
As the field progresses, researchers are increasingly applying LVLMs to a wider range of visual perception tasks, such as visual grounding \citep{peng2024kosmos} and reasoning segmentation \citep{lai2024lisa, liu2025segzero}
, often incorporating task-specific modules or techniques.

Through an analysis of diverse visual perception tasks, we observe that many can be categorized into three fundamental types:
detection (e.g., object detection \citep{lin2014mscoco}, visual grounding \citep{yu2016refcoco}), segmentation (e.g., referring expression segmentation \citep{yu2016refcoco}, reasoning segmentation \citep{lai2024lisa}),
and counting (e.g., object counting \citep{paiss2023countbench}).
Notably, our analysis reveals that these three task types share a common structure as multi-object cognition problems, suggesting that they can be addressed through a unified framework.

Moreover, recent studies have explored the integration of reinforcement learning (RL) into LVLMs \citep{r1v,liu2025visualrft,liu2025segzero,zheng2025easyr1}.
Works such as VisualRFT \citep{liu2025visualrft} and Seg-Zero \citep{liu2025segzero} demonstrate that RL can enhance reasoning in visual perception tasks.
However, these approaches often employ RL in a task-specific manner, training with different data for different tasks, which may limit their scalability and generalizability.

Building on these insights, we propose {\ours}, a unified framework that addresses diverse visual perception tasks through a shared architecture.
The framework's core capabilities, which include advanced reasoning and multi-object cognition, are enabled through RL and a unified reward mechanism.
Format rewards, including thinking rewards that promote structured reasoning and non-repeat rewards that prevent redundant reasoning patterns.
Accuracy rewards, comprising multi-object IoU rewards and L1 rewards for precise localization, strengthen multi-object cognition. Unlike previous approaches like Kosmos \citep{peng2024kosmos} that use cross-entropy loss, our RL framework requires optimal prediction-to-ground-truth matching. We address this challenge by implementing an efficient matching pipeline combining the batch computing and the Hungarian algorithm, significantly improving computational efficiency while maintaining matching accuracy. 

\begin{figure}[t]
  \centering
   \includegraphics[width=1.0\linewidth]{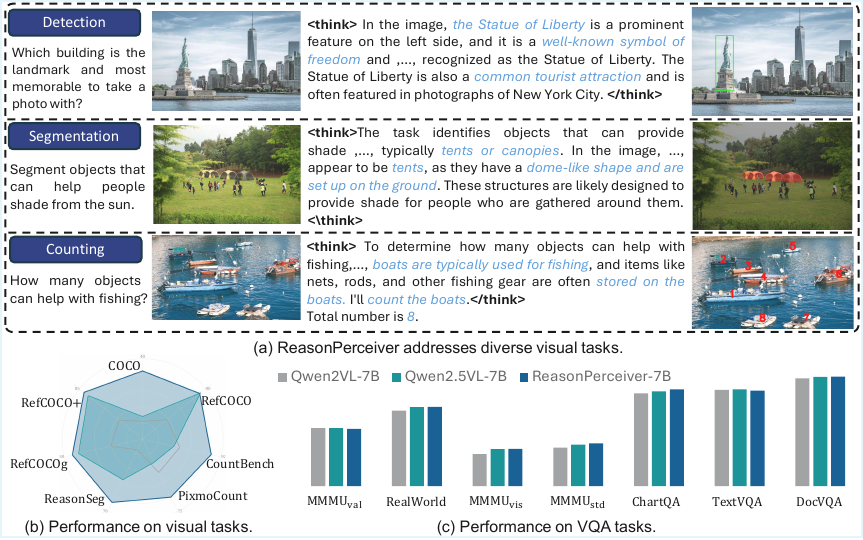}
   \caption{(a) {\ours} addresses diverse tasks within a unified framework. It generates a reasoning process and outputs the expected result corresponding to each query. (b) {\ours} significantly outperforms Qwen2.5VL. (c) {\ours} retains strong VQA capabilities.}
   \label{fig:teaser}
\end{figure}

To comprehensively evaluate model performance, we conduct extensive experiments with {\ours} across 10 diverse tasks spanning three fundamental types: detection, segmentation, and counting.
Remarkably, our {\ours}-7B model achieves strong performance despite being trained on only 7k samples, demonstrating both robust test-time reasoning capabilities and effective multi-task generalization, as shown in \Cref{fig:teaser} (a)-(b).
Experimental results show significant improvements over baseline models, with relative gains of 29.1\% on COCO-val (detection), 22.1\% on ReasonSeg-test (segmentation), and 13.2\% on CountBench-test (counting), validating the effectiveness of our unified approach.
Additionally, {\ours} exhibits visual question answering capabilities comparable to state-of-the-art models, as shown in \Cref{fig:teaser} (c). Human evaluation also indicates {\ours} generates faithful and reliable reasoning process even without training on annotated reasoning data.

Our contributions are summarized as follows:
\begin{itemize}
    \item We propose {\ours}, a unified framework for visual perception tasks. Through carefully crafted rewards and training strategy, {\ours} has strong multi-task capability, addressing diverse visual perception tasks within a shared model.
    \item Experimental results show that {\ours} achieves superior performance across ten diverse visual perception tasks within a single unified framework, outperforming baseline models by a significant margin.
    \item Through extensive ablation studies, we validate the effectiveness of our design and offer critical insights into the application of RL in LVLMs.
\end{itemize}

\section{Related Works}
\subsection{Large Vision-language Models}
Following LLaVA's \citep{liu2023llava} pioneering work on visual instruction tuning for large vision-language models,
subsequent studies \citep{wang2024qwen2vl,llama32,openai2025gpt41,bai2025qwen25vl,li2024mgm,zhong2024lyra} have adopted this paradigm for vision-language conversation.
Beyond visual conversation tasks, LVLMs have been extended to diverse vision applications, including visual grounding \citep{peng2024kosmos} and reasoning segmentation \citep{lai2024lisa}.
Notely, the recent GPT-4.1 \citep{openai2025gpt41} demonstrates state-of-the-art performance in multi-modal information processing and visual reasoning.
Although these models are evaluated on specific tasks, their performance has not been systematically evaluated under a unified visual perception framework.

\subsection{Reinforcement Learning in Large Models}
In the field of large language model (LLMs), various reinforcement learning (RL) algorithms are used to enhance model performance,
such as reinforcment learning from human feedback (RLHF) \citep{ouyang2022instructgpt}, direct preference optimization (DPO) \citep{rafailov2023dpo} and proximal policy optimization (PPO) \citep{schulman2017ppo}.
The recent DeepSeek R1 \citep{guo2025deepseekr1}, trained using Group Relative Policy Optimization (GRPO) \citep{shao2024deepseekmath}, demonstrates remarkable test-time scaling capabilities, significantly improving reasoning ability and overall performance. 
Building on these advances, researchers try to apply these RL techniques to LVLMs.
Notable efforts include Visual-RFT \citep{liu2025visualrft}, EasyR1 \citep{zheng2025easyr1} and Seg-Zero\citep{liu2025segzero}, all of which exhibit strong reasoning capabilities and achieve impressive performance.

\section{Method}
\label{sec:method}

To develop a unified visual perception model capable of solving diverse vision tasks, we identify and analyze the representative visual perception tasks, then reformulate their inputs and outputs into a set of three fundamental task categories (\Cref{sec:task_category}). Next, we detail the architecture of our {\ours} model (\Cref{sec:vision_one}). Additionally, we present the unified reward mechanism employed for training our model (\Cref{sec:rewards}). Finally, we elaborate on our training strategy of multi-object cognition (\Cref{sec:training}).

\subsection{Preliminary}
\label{sec:preliminary}

\begin{wrapfigure}{t}{0.4\textwidth}
    \centering
    \vspace{-1.5em}
    \includegraphics[width=1.\linewidth]{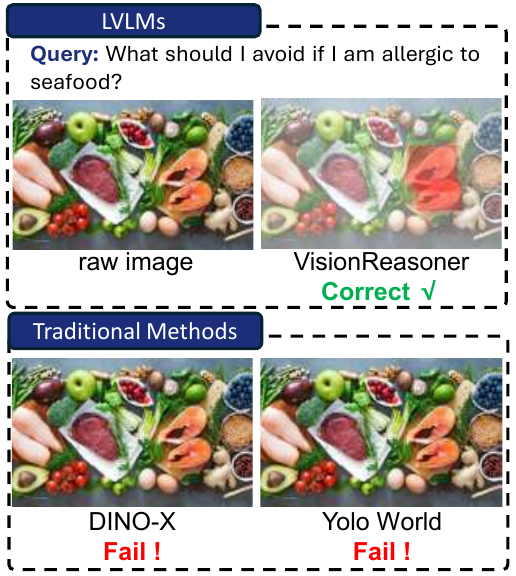}
    \caption{{\ours} correctly localizes objects from a complex instruction, whereas both commercial DINO-X and open-source YOLO-World fail.}
    \vspace{-1em}
    \label{fig:compare_traditional_methods}
\end{wrapfigure}

\textbf{Traditional Vision Methods vs. LVLMs.} Although traditional vision models \citep{cheng2024yoloworld, ren2024dinox} achieve strong performance on standard visual perception benchmarks \citep{lin2014mscoco}, they are inherently limited to processing simple categorical queries and struggle with complex, compositional, or reasoning-intensive instructions. In contrast, LVLMs can interpret and respond to nuanced, open-ended queries. As illustrated in \Cref{fig:compare_traditional_methods}, {\ours} successfully localizes and identifies target objects that traditional approaches fail to detect, highlighting the necessity to integrate LVLMs into visual perception pipelines where reasoning are essential.

\textbf{Group Relative Policy Optimization (GRPO).} The GRPO is a on-policy reinforcement learning algorithm. For each input $x$, the old policy model $\pi_{\theta_{old}}$ from previous step generate a group of rollouts $\{o_{i}\}^G_{i=1}$. Then reward functions are used to calculate rewards for each $o_i$, getting $\{r_{i}\}^G_{i=1}$. We design a unified reward mechanism and the relative advantage is calculated as:
\begin{equation}
\footnotesize
    A_i = \frac{r_i - \text{mean}(\{r_1, r_2, \dots, r_G\})}{\text{std}(\{r_1, r_2, \dots, r_G\})}.
\end{equation}

The GRPO maximizes the following objective and
optimizes the model $\pi_\theta$:

\begin{equation}
\label{eq:objective}
\footnotesize
\begin{split}
&\mathcal{J}_{\text{GRPO}}(\theta) = \mathbb{E}_{x \sim \text{Train Batch},\, \{o_i\}_{i=1}^G \sim \pi_{\theta_{\text{old}}}(O|x)} \\
&\quad \left[ \frac{1}{G} \sum_{i=1}^G \min\left( \frac{\pi_\theta(o_i\mid x)}{\pi_{\theta_{\text{old}}}(o_i\mid x)} A_i,\, \text{clip}\left( \frac{\pi_\theta(o_i\mid x)}{\pi_{\theta_{\text{old}}}(o_i\mid x)}, 1-\varepsilon, 1+\varepsilon \right) A_i \right) - \beta D_{\text{KL}}(\pi_\theta \parallel \pi_{\text{ref}}) \right].
\end{split}
\end{equation}

\subsection{Task Reformulation and Categorization}  
\label{sec:task_category}

Our analysis of vision perception tasks \citep{yu2016refcoco,lin2014mscoco,lai2024lisa,deitke2024molmo} reveals that many of them can be categorized into three fundamental task types. 
Here we take ten visual perception tasks for illustration. Further details are provided in the \Cref{appdx:task_reformulation}. 

\textbf{Detection.} Given an image $\mathbf{I}$ and a text query $\mathbf{T}$, the detection task type aims to generate a set of bounding boxes $\{\mathbf{B}_i\}_{i=1}^{N}$ that localize objects of interest. This type requires multi-object cognition ability. This category includes tasks such as Visual Grounding \citep{yu2016refcoco,kazemzadeh2014referitgame} and Object Detection \citep{lin2014mscoco}. 

\textbf{Segmentation.} Given an image $\mathbf{I}$ and a text query $\mathbf{T}$, the segmentation task type aims to generate a set of binary segmentation masks $\{\mathbf{M}_i\}_{i=1}^{N}$ that identify the regions of interest. We address this type by detect-then-segment paradigm. This category includes tasks such as Referring Expression Segmentation \citep{kazemzadeh2014referitgame,yu2016refcoco} and Reasoning Segmentation \citep{lai2024lisa,yang2023lisa++}. 

\textbf{Counting.} Given an image $\mathbf{I}$ and a text query $\mathbf{T}$, the counting task type aims to estimate the number of target objects specified by the query.  We address this type by detect-then-count paradigm. This category includes tasks such as Object Counting \citep{deitke2024molmo,paiss2023countbench}.  


\begin{figure}[t]
  \centering
   \includegraphics[width=1.0\linewidth]{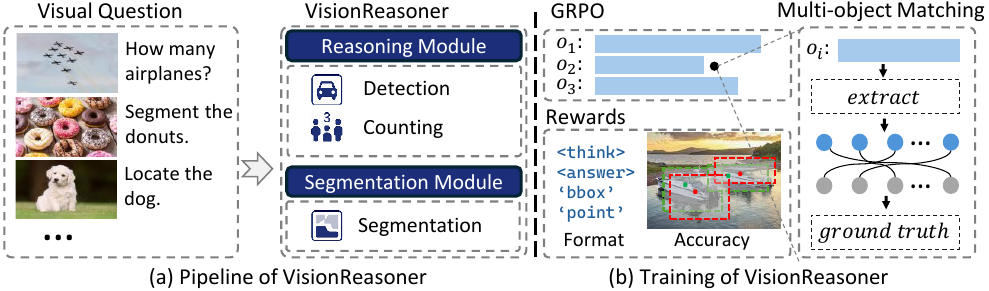}
   \caption{Illustration of {\ours}. (a) For a given image $\mathbf{I}$ and text instruction $\mathbf{T}$, our model generates the expected output corresponding to the instruction. (b) For each observation $o_i$, we calculate the rewards (\Cref{sec:rewards}) and attain the optimal match of multi-objects (\Cref{sec:training})}
   \label{fig:vision_one}
\end{figure}

\subsection{{\ours} Model}
\label{sec:vision_one}
Our {\ours} $\mathcal{F}$ model incorporates a reasoning module, which processing image and locates targeted objects, and a segmentation module that produces segmentation masks if needed. The whole architecture is shown in \Cref{fig:vision_one} (a). 
The key to $\mathcal{F}$ lies in its multi-object cognition capabilities, which is critical to enables {\ours} to address three fundamental task types: detection, segmentation, counting. 
Specifically, given an image $\mathbf{I}$, a text query $\mathbf{T}$, the {\ours} $\mathcal{F}$ generates an interpretable reasoning process, and then produces the bounding boxes $\{\mathbf{B}_i\}_{i=1}^{N}$ and central points $\{\mathbf{P}_i\}_{i=1}^{N}$ of targeted objects corresponding to $\mathbf{T}$. Then $\{\mathbf{B}_i\}_{i=1}^{N}$ and $\{\mathbf{P}_i\}_{i=1}^{N}$ serve as bridge to connect the segmentation module, producing binary masks $\{\mathbf{M}_i\}_{i=1}^{N}$ if needed. 
This process can be formulated as:
\begin{equation}
    (\{\mathbf{B}_i, \mathbf{M}_i\})_{i=1}^{N} = \mathcal{F}(\mathbf{I}, \mathbf{T}).
\end{equation}

During inference, the user provide the input image $\mathbf{I}$ and text prompt $\mathbf{T}$, and define a specified task type $\mathbf{C} \in \{\text{detection, segmentation, counting}\}$. The system then produces the expected outputs as follows:
\begin{equation}
    \text{Output} = 
    \begin{cases}
        \{\mathbf{B}_i\}_{i=1}^{N}, & \text{if } \mathbf{C} \text{ is detection},  \\
        \{\mathbf{M}_i\}_{i=1}^{N}, & \text{if }\mathbf{C} \text{ is segmentation},  \\
        N,                        & \text{if } \mathbf{C} \text{ is counting}.   \\  
    \end{cases}
\end{equation}
In this way, our {\ours}  can process diverse perception tasks in a unified manner within a shared framework. Moreover, our framework can be easily extended to other visual perception tasks as illustrated in \Cref{appdx:extension}.

\subsection{Unified Reward Mechanism}
\label{sec:rewards}
As illustrated in \Cref{sec:preliminary}, the core in group relative RL is the design of rewards. We design a unified reward mechanism for visual perception tasks, including format rewards and accuracy rewards. We use target object bboxes and center points to calculate the rewards rather than binary masks. These rewards jointly guide the optimization process by reinforcing both structural correctness and multi-object recognition performance. The model is capable of addressing diverse visual perception tasks after training within this unified reward mechanism. The total reward is the sum of all rewards.

\textbf{Thinking Format Reward.} This reward is 1.0 if the model output a thinking process between \textless think\textgreater ~and~  \textless /think\textgreater tags, and output the final answer between the \textless answer\textgreater ~and~ \textless /answer\textgreater tags. 

\textbf{Answer Format Reward.} We use bounding boxes $\{\mathbf{B}_i\}_{i=1}^{N}$ and points $\{\mathbf{P}_i\}_{i=1}^{N}$ as the answer as it has better training efficiency. So this reward restrict the model output answer in $[\{'\text{bbox\_{2d}}': [x_1, y_1, x_2, y_2], '\text{point\_2d}': [x_1, y_1]\}, ...].$ The reward is 1.0 if correct else 0.0.

\textbf{Non Repeat Format Reward.} We split the reasoning process into sentences to detect repeated pattern. A reward of 1.0 is assigned for those with unique or non-repetitive thinking processes. 

\textbf{Bboxes IoU Reward.} Given a set of $N$ ground-truth bounding boxes and $K$ predicted bounding boxes, this reward computes their optimal one-to-one matched Intersection-over-Union (IoU) scores.
For each IoU exceeding 0.5, we increment the reward by $ \cfrac{1}{\max\{N, K\}}$.

\textbf{Bboxes L1 Reward.} Given a set of $N$ ground-truth bounding boxes and $K$ predicted bounding boxes, this reward computes their one-to-one matched L1 distance. For each L1 distance below the threshold of 10 pixel, we increment the reward by $ \cfrac{1}{\max\{N, K\}}$.

\textbf{Points L1 Reward.} Given a set of $N$ ground-truth points and $K$ predicted points, this reward computes their one-to-one matched L1 distance. For each L1 distance below the threshold of 30 pixel, we increment the reward by $ \cfrac{1}{\max\{N, K\}}$.

\subsection{Multi-object Cognition in LVLMs} 
\label{sec:training} 


Unlike the auto-regressive training paradigm \citep{peng2024kosmos,bai2025qwen25vl} in supervised fine-tuning, RL framework requires optimal prediction-to-ground-truth matching for reward calculation. To address this, we derive the necessary data and implement an effective matching strategy.

\begin{wrapfigure}{r}{0.6\linewidth} 
\vspace{-1em} 
\begin{minipage}{\linewidth}
\begin{algorithm}[H]
\footnotesize
\caption{Multi-object Matching}
\label{alg:matching}
\DontPrintSemicolon
\SetAlgoLined
\KwIn{pred bboxes $\bm{b}_{\text{pred}}\in\mathbb{R}^{K\times4}$; pred points $\bm{p}_{\text{pred}}\in\mathbb{R}^{K\times2}$; GT bboxes $\bm{b}_{\text{gt}}\in\mathbb{R}^{N\times4}$; GT points $\bm{p}_{\text{gt}}\in\mathbb{R}^{N\times2}$}
\SetKwFunction{Hungarian}{Hungarian}
\SetKwProg{Fn}{Function}{:}{}
\Fn{AccuracyReward($\bm{b}_{\text{pred}}, \bm{p}_{\text{pred}}, \bm{b}_{\text{gt}}, \bm{p}_{\text{gt}}$)}{
    $r \gets 0$; $L_{\max} \gets \max(K,N)$;
    
    $IoU \leftarrow \text{BatchIoU}(\bm{b}_{\text{pred}}, \bm{b}_{\text{gt}}) \in \mathbb{R}^{K\times N}$\;
    $BL1 \leftarrow \text{BatchBoxL1Distance}(\bm{b}_{\text{pred}}, \bm{b}_{\text{gt}}) \in \mathbb{R}^{K\times N}$\;
    $PL1 \leftarrow \text{BatchPointL1Distance}(\bm{p}_{\text{pred}}, \bm{p}_{\text{gt}}) \in \mathbb{R}^{K\times N}$\;

    $R_{\text{IoU}} \leftarrow [IoU>\text{IoU threshold}]$\; 
    $R_{\text{BL1}} \leftarrow [BL1<\text{Box L1 threshold}]$\; 
    $R_{\text{PL1}} \leftarrow [PL1<\text{Point L1 threshold}]$\;
    $C \leftarrow (3 - (R_{\text{IoU}} + R_{\text{BL1}} + R_{\text{PL1}}) ) \in \mathbb{R}^{K\times N}$\;

    $(\bm{r},\bm{c}) \leftarrow \Hungarian(C)$\;
    $\text{total} \leftarrow 3|\bm{r}| - \sum_t C_{\bm{r}_t,\bm{c}_t}$\;

    $r \leftarrow \text{total}/L_{\max}$; \Return $r$\;
}
\KwOut{Accuracy reward $r$}
\end{algorithm}
\end{minipage}
\vspace{-2em} 
\end{wrapfigure}

\textbf{Multi-object Data Preparation.} We derive bboxes and points directly from the original mask annotations in existing segmentation datasets (e.g., RefCOCOg \citep{yu2016refcoco}, LISA++\citep{yang2023lisa++}). Specifically, for a given binary mask of an object, we determine its bounding box by extracting the leftmost, topmost, rightmost, and bottommost pixel coordinates. Additionally, we compute the center point coordinates of the mask. We process multiple objects per image by: (i) using one central point (ii) joining all textual descriptions with the conjunction `and', and (iii) concatenating all associated bounding boxes and center points into list per image.

\textbf{Multi-object Matching.} Our framework addresses multi-object matching through batch computation and the Hungarian algorithm, which optimally solves the many-to-many matching problem for bounding boxes IoU rewards, bounding boxes L1 rewards, and points L1 rewards. As shown in \Cref{fig:vision_one} (b), for each observation $o_j$, which contains a list of bboxes $\{\mathbf{B_{pred}}_i\}_{i=1}^{K}$ and points $\{\mathbf{P_{pred}}_i\}_{i=1}^{K}$, we calculate its reward scores with the ground-truth bboxes $\{\mathbf{B_{GT}}_i\}_{i=1}^{N}$ and points $\{\mathbf{P_{GT}}_i\}_{i=1}^{N}$ by implementing batch computation. We then calculate the optimal one-to-one matching with using Hungarian algorithm. The pseudocode of multi-object matching is shown in \Cref{alg:matching}. These design guarantees optimal assignment between predictions and ground truth annotations while achieving high computational efficiency.

    




\section{Experiments}
\label{sec:experiments}

\subsection{Experimental Settings}
\textbf{Evaluation Benchmark.} We use ten benchmarks to evaluate model performance across general vision perception tasks, including three fundamental task types: detection, segmentation and counting. Specifially, we employ COCO \citep{lin2014mscoco} and RefCOCO(+/g) \citep{yu2016refcoco} for detection evaluation; RefCOCO(+/g) and ReasonSeg \citep{lai2024lisa} for segmentation evaluation; PixMo-Count \citep{deitke2024molmo} and CountBench \citep{paiss2023countbench} for counting evaluation. Details of benchmarks and metrics can be found in \Cref{appdx:detail_benchmark}.

\textbf{Training Data.} The training data is sourced from the training splits of four datasets: LVIS \citep{gupta2019lvis}, RefCOCOg \citep{yu2016refcoco}, gRefCOCO \citep{liu2023grefcoco}, and LISA++ \citep{yang2023lisa++}.
We randomly collect approximately 7k training samples, with around 1,800 from each dataset.



\textbf{Implementation Details.} We initialize {\ours} with Qwen2.5-VL and SAM2. 
We employ a batch size of 16 and a learning rate of 1e-6. The training objective is \Cref{eq:objective}. 

\setlength{\tabcolsep}{8.5pt}{
\begin{table}[t]
\centering
\caption{Performance comparison on detection tasks.}
\footnotesize
\begin{tabular}{l|c|cc|cc|cc|c}
\toprule
\multirow{4}{*}{\textbf{Method}} & \multicolumn{7}{c}{\textbf{Detection}} \vline & \multirow{4}{*}{\textbf{Avg.}} \\
\cmidrule{2-8} 
    & \multicolumn{1}{c}{COCO} \vline & \multicolumn{2}{c}{RefCOCO} \vline & \multicolumn{2}{c}{RefCOCO+} \vline & \multicolumn{2}{c}{RefCOCOg} \vline  \\
\cmidrule{2-8} 
    & val  & val & testA &  val & testA & val & test \\
\midrule
\multicolumn{4}{l}{\textcolor{gray}{Task-specific Models}} \\
VGTR & - & 79.0 & 82.3 & 63.9 & 70.1  & 65.7 & 67.2 & -\\
TransVG & - & 81.0 & 82.7  & 64.8 & 70.7  & 68.7 & 67.7 & - \\
RefTR & - & 85.7 & 88.7  & 77.6 & 82.3  & 79.3 & 80.0 & - \\
MDETR & - & 86.8 & 89.6  & 79.5 & 84.1  & 81.6 & 80.9 & - \\
OWL-ViT & 30.9 & - & - & - & - & - & - & -\\
YOLO-World-S & 37.6  & - & - & - & - & - & - & - \\ 
GLIP-T & 46.6 & 50.4 & 54.3  & 49.5 & 52.8  & 66.1 & 66.9 & 55.2 \\ 
G-DINO-T & 48.4 & 74.0 & 74.9 & 66.8 & 69.9 & 71.1 & 72.1 & 68.2\\ 
DQ-DETR & 50.2 & 88.6 & 91.0 & 81.7 & 86.2  & 82.8 & 83.4 & \textbf{80.6} \\ 
\midrule
\multicolumn{4}{l}{\textcolor{gray}{Large Vision-language Models}} \\
Shikra-7B & - & 87.0 & 90.6 & 81.6 & 87.4 & 82.3 & 82.2 & - \\
InternVL2-8B & - & 87.1 & 91.1 & 79.8 & 87.9 & 82.7 & 82.7 & - \\
Qwen2-VL-7B  & 28.3 & 80.8 & 83.9 & 72.5 & 76.5 & 77.3 & 78.2 & 71.1 \\
Qwen2.5-VL-7B  & 29.2 & 88.8 & 91.7 & 82.3 & 88.2 & 84.7 & 85.7 & 78.6 \\
\cellcolor[HTML]{efefef}{{\ours}-7B}  & \cellcolor[HTML]{efefef}{37.7} & \cellcolor[HTML]{efefef}{88.6} & \cellcolor[HTML]{efefef}{90.6} & \cellcolor[HTML]{efefef}{83.6} & \cellcolor[HTML]{efefef}{87.9} & \cellcolor[HTML]{efefef}{86.1} & \cellcolor[HTML]{efefef}{87.5} & \cellcolor[HTML]{efefef}{\textbf{80.3}} \\
\bottomrule
\end{tabular}
\label{table:performance_detection}
\end{table}}

\setlength{\tabcolsep}{4.5pt}{
\begin{table}[t]
\centering
\caption{Performance comparison on segmentation tasks and counting tasks. We use SAM2 for vision-language models if necessary in segmentation tasks. }
\footnotesize
\begin{tabular}{l|cc|c|c|c|c|cc|c|c}
\toprule
\multirow{4}{*}{\textbf{Method}} & \multicolumn{5}{c}{\textbf{Segmentation}} \vline & \multirow{3}{*}{\textbf{Avg.}} & \multicolumn{3}{c}{\textbf{Counting}} \vline & \multirow{4}{*}{\textbf{Avg.}}  \\
\cmidrule{2-6} \cmidrule{8-10} 
    & \multicolumn{2}{c}{ReasonSeg} \vline & \multicolumn{1}{c}{RCO} \vline & \multicolumn{1}{c}{RCO+} \vline & \multicolumn{1}{c}{RCOg} \vline & & \multicolumn{2}{c}{Pixmo} \vline & \multicolumn{1}{c}{Count} \vline  \\
\cmidrule{2-6} \cmidrule{8-10} 
    & val & test  & testA  & testA & test & & val & test & test\\
\midrule
\multicolumn{4}{l}{\textcolor{gray}{Task-specific Models}} \\
LAVT & - & - & 75.8 & 68.4 & 62.1 & - & - & - & - & - \\
ReLA & 22.4 & 21.3 & 76.5 & 71.0 & 66.0 & 51.4 & - & - & - & - \\
\midrule
\multicolumn{4}{l}{\textcolor{gray}{Large Vision-language Models}} \\
LISA-7B & 44.4 & 36.8 & 76.5 & 67.4 & 68.5 & 58.7 & - & - & - & -\\
LLaVA-OV-7B & - & - & - & - & - & - & 55.8 & 53.7 & 78.8 & 62.8 \\
GLaMM-7B & - & - & 58.1 & 47.1 & 55.6 & - & - & - & - & \\  
PixelLM-7B & - & - & 76.5& 71.7 & 70.5 & - & - & - & - & \\
Seg-Zero-7B & 62.6 & 57.5 & 80.3 & 76.2 & 72.6 & 69.8 & - & - & - &\\
Qwen2-VL-7B & 44.5 & 38.7  & 68.7  & 65.7 & 63.5 & 56.2 & 61.6 & 56.3 & 80.4 & 66.1 \\ 
Qwen2.5-VL-7B & 56.9 & 52.1  & 79.9  & 76.8 & 72.8 & 67.7 & 58.1 & 53.1 & 78.8 & 63.6 \\ 
\cellcolor[HTML]{efefef}{{\ours}-7B}  & \cellcolor[HTML]{efefef}{66.3} & \cellcolor[HTML]{efefef}{63.6} & \cellcolor[HTML]{efefef}{78.9}  & \cellcolor[HTML]{efefef}{74.9}  & \cellcolor[HTML]{efefef}{71.3} & \cellcolor[HTML]{efefef}{\textbf{71.0}} & \cellcolor[HTML]{efefef}{70.1} & \cellcolor[HTML]{efefef}{70.7} & \cellcolor[HTML]{efefef}{89.2} & \cellcolor[HTML]{efefef}{\textbf{76.7}} \\ 
\bottomrule
\end{tabular}
\label{table:performance_segmentation_count}
\end{table}}

\subsection{Main Results}
We compare the results with LVLMs and task-specific models on each of the three fundamental task types. It is worthy note that our {\ours} is capable of handling different tasks within the same model and is evaluated in a zero-shot manner. 

\textbf{Detection.} We compare {\ours} with several state-of-the-art LVLMs, including Shikra \citep{chen2023shikra}, InternVL2-8B \citep{internvl2},  Qwen2-VL-7B \citep{wang2024qwen2vl} and Qwen2.5VL-7B \citep{bai2025qwen25vl}. For task-specific models, we evaluate against VGTR \citep{da2023vgtr}, TransVG \citep{deng2021transvg}, RefTR \citep{li2021reftr}, MDETR \citep{kamath2021mdetr}, OWL-ViT \citep{minderer2022owlvit}, YOLO-World \citep{cheng2024yoloworld}, GroundingDINO \citep{liu2024grounding}, DQ-DETR \citep{liu2023dqdetr}, GLIP \citep{li2022glip}. Since LVLMs do not output confidence score, we approximate it using the ratio of the bounding box area to the total image area (bbox\_area / image\_area) to enable compatibility with COCOAPI \citep{cocoapi}. However, this coarse approximation leads to underestimated AP scores. As shown in \Cref{table:performance_detection}, {\ours} achieves superior performance among LVLMs. While our model shows a performance gap compared to some task-specific baselines on COCO datasets, it maintains competitive advantages due to its superior generalization capability.

\textbf{Segmentation.} We evaluate {\ours} against state-of-the-art LVLMs, including LISA \citep{lai2024lisa}, GLaMM \citep{rasheed2024glamm}, PixelLM \citep{ren2024pixellm}, Seg-Zero \citep{liu2025segzero}, Qwen2-VL \citep{wang2024qwen2vl} and Qwen2.5VL \citep{bai2025qwen25vl}. For these LVLMs, we first extract bounding box predictions and subsequently send them into SAM2\citep{ravi2024sam2} to generate segmentation masks. We also compare task-specific models, including LAVT \citep{yang2022lavt} and ReLA \citep{liu2023gres}. For models that do not report gIoU, we report their cIoU as an alternative. As shown in \Cref{table:performance_segmentation_count}, {\ours} achieves state-of-the-art performance, outperforming both general-purpose LVLMs and task-specific approaches.

\textbf{Counting.}  We evaluate {\ours} against state-of-the-art LVLMs, including LLaVA-OneVision \citep{li2024llavaov}, Qwen2-VL-7B \citep{wang2024qwen2vl} and Qwen2.5VL-7B \citep{bai2025qwen25vl}. We evaluate these LVLMs in a first-detect-then-count manner. As shown in \Cref{table:performance_segmentation_count}, {\ours} achieves state-of-the-art performance. 

\begin{table}[t]
  \begin{minipage}[t]{0.3\textwidth}
    \centering 
    \caption{Comparison of multi-object matching. Our code achieves a $4 \times$ speedup.}
    \vspace{-2.5pt}
    \footnotesize
    \label{tab:multiobject_match}
    \setlength{\tabcolsep}{1pt}{
    \begin{tabular}{cc|c}
      \toprule
      \textbf{Hungarian} & \textbf{BatchComp} &  \textbf{Time (s)} \\
      \midrule
      \checkmark   &              &  $2 \times 10^{-3}$ \\
      \checkmark   & \checkmark   &  $5 \times 10^{-4}$ \\
      \bottomrule
    \end{tabular}}
  \end{minipage}
  \hfill
  \begin{minipage}[t]{0.3\textwidth}
    \centering
    \caption{Comparison on the reasoning length. }
    \footnotesize
    \label{tab:reason_len}
    \setlength{\tabcolsep}{1pt}{
    \begin{tabular}{l|c}
      \toprule
      \textbf{Data} & \textbf{Avg. Len (\# words)} \\
      \midrule
             COCO      & 62 \\
             RefCOCOg  & 65 \\ 
             ReasonSeg  & 71 \\ 
      \bottomrule
    \end{tabular}}
  \end{minipage}
  \hfill
  \begin{minipage}[t]{0.3\textwidth}
    \centering
    \caption{Comparison on different RL algorithm.}
    \footnotesize
    \label{tab:diff_rl}
    \setlength{\tabcolsep}{8pt}{
    \begin{tabular}{l|c}
      \toprule
      \textbf{RL} & \textbf{ReasonSeg-val} \\
      \midrule
             Baseline  & 56.9 \\
             GRPO      & 61.9 \\
             DAPO      & 61.7 \\ 
      \bottomrule
    \end{tabular}}
  \end{minipage}
\end{table}





\begin{figure}[t]
    \centering
    \includegraphics[width=1.\linewidth]{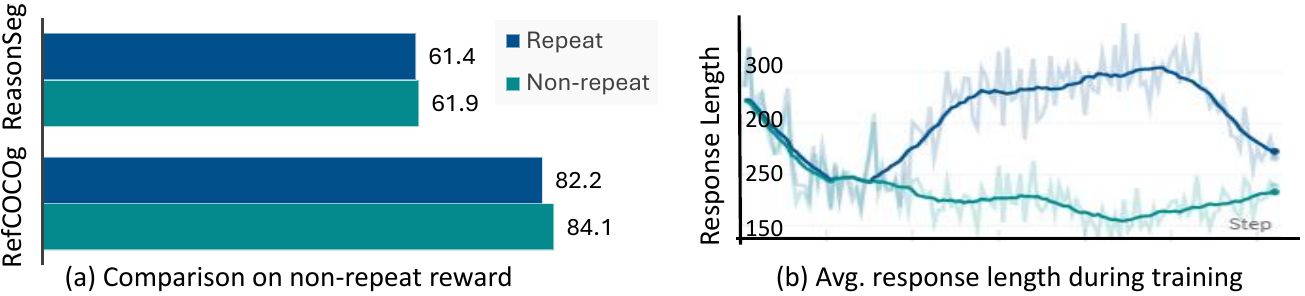}
    \caption{Ablation on non-repeat reward. (a) Consistent performance gain across different datasets using non-repeated reward. (b) Non-repeat rewards lead to shorter response lengths.}
    \label{fig:non_repeat}
\end{figure}

\begin{figure}[t]
  \centering 
  \begin{minipage}[b]{\linewidth}
    \centering
    \captionof{table}{Performance comparison on different training data.}
    \footnotesize
    \setlength{\tabcolsep}{3pt}{
    \begin{tabular}{l|cccc|c|c|c}
    \toprule
    \multirow{2}{*}{\textbf{Method}} & \multicolumn{4}{c}{\textbf{Training Data}} \vline & \multicolumn{1}{c}{\textbf{Det}} \vline & \multicolumn{1}{c}{\textbf{Seg}} \vline & \multirow{2}{*}{\textbf{Avg.}}\\
    \cmidrule{2-7}
         & RefCOCOg & gRefCOCO & LVIS & LISA++ & RefCOCOg-val & ReasonSeg-val  \\
    \cmidrule{1-8}
    
    \multirow{4}{*}{{\ours}-7B}   & \checkmark &  &  &  & 84.1 & 61.9 & 73.0 \\ 
                                  & \checkmark & \checkmark  &  &  & 85.8 & 63.8 & 74.8 \\
                                  & \checkmark & \checkmark  & \checkmark  &  &  85.5 & 64.2 & 74.9\\
                                  & \checkmark & \checkmark  & \checkmark  & \checkmark & \textbf{86.1}  & \textbf{66.3} & \textbf{76.2}\\
    \bottomrule
    \end{tabular}}
    \label{table:performance_train_data}
  \end{minipage}

  \vspace{1.em}
  
  \begin{minipage}[b]{\linewidth}
    \centering
    \captionof{table}{Performance comparison on VQA tasks.}
    \footnotesize
    \setlength{\tabcolsep}{2pt}{
    \begin{tabular}{l|cccccc}
    \toprule
    \textbf{Method}  & OCRBench & RealworldQA  & MMMUPro$_{vision}$ & MMMUPro$_{std}$ & ChartQA & DocVQA\\
    \midrule
    Qwen2VL-7B   & 809 & 66.1 & 28.0 & 33.8 & 81.4 & 94.5 \\
    Qwen2.5VL-7B & 822 & 69.2 & 32.4 & 36.4 & 83.1 & 95.7\\
    \cellcolor[HTML]{efefef}{{\ours}-7B}    & \cellcolor[HTML]{efefef}{\textbf{825}} & \cellcolor[HTML]{efefef}{\textbf{69.5}} & \cellcolor[HTML]{efefef}{\textbf{32.6}} & \cellcolor[HTML]{efefef}{\textbf{37.4}} & \cellcolor[HTML]{efefef}{\textbf{84.9}} & \cellcolor[HTML]{efefef}{\textbf{96.0}} \\
    \bottomrule
    \end{tabular}}
    \label{table:vqa_results}
  \end{minipage}
\end{figure}




\begin{figure}[t]
  \centering
  \begin{minipage}[b]{0.43\linewidth}
    \centering
    \includegraphics[width=\linewidth]{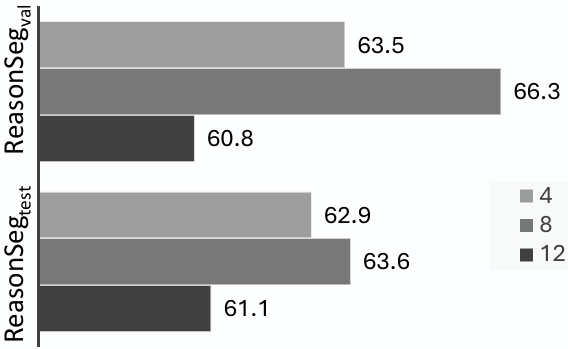}
    \captionof{figure}{Different sampling number.}
    \label{fig:sampling_num}
  \end{minipage}
  \hfill
  \begin{minipage}[b]{0.47\linewidth}
    \centering
    \includegraphics[width=\linewidth]{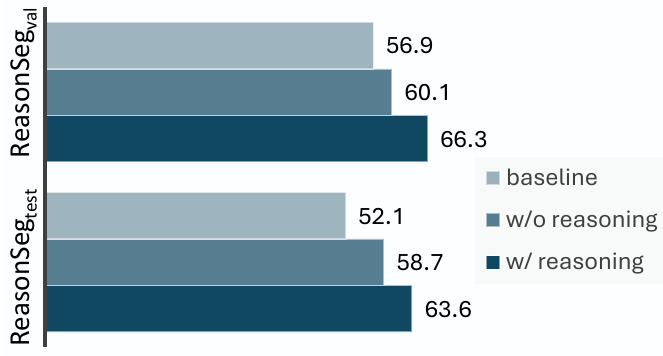}
    \captionof{figure}{Reasoning vs.~no reasoning}
    \label{fig:reasoning}
  \end{minipage}
  
  
\end{figure}





\subsection{Ablation Study}
We perform ablation studies to assess the effectiveness of our design and validate the optimal hyperparameter selection and training recipe design for {\ours}. 
We also evaluate {\ours} on VQA tasks.

\textbf{Multi-object Matching.}
We quantitatively assess the efficiency of our two key design choices for multi-object matching: the Hungarian algorithm and batch computation. In a scenario with 30 objects, \Cref{tab:multiobject_match} demonstrates that a non-batch matching require $2\times10^{-3}$ seconds to complete, while our optimized approach achieves matching in just $5\times10^{-4}$ seconds - a $4 \times$ speedup.

\textbf{Reasoning Length.} As shown in \Cref{tab:reason_len}, our analysis reveals that the model's reasoning length adapts dynamically to text query complexity. Specifically, for simple class names in COCO and short phrases in RefCOCOg, the reasoning process is relatively concise. In contrast, complex reasoning-intensive queries in ReasonSeg require longer reasoning processes.

\textbf{Non Repeat Reward.}
\Cref{fig:non_repeat} (a) presents the performance comparison with and without the non-repeat reward. Models are trained only on 2,000 samples from RefCOCOg. The model achieves better results when trained using the non-repeat reward. Additionally, model without non-repeat reward tends to generate longer reasoning processes, as shown in \Cref{fig:non_repeat} (b), and we observe repetitive reasoning patterns during inference. 

\textbf{Different RL Algorithm.} We use different on-policy RL training algorithm: the GRPO \citep{shao2024deepseekmath} and DAPO \citep{yu2025dapo}. Models are trained only on 2,000 samples from RefCOCOg. As shown in \Cref{tab:diff_rl}, performance consistently improves across both algorithms, demonstrating that our training framework is both stable and generalizable.

\textbf{Different Training Data.}
We conduct an ablation study on different training datasets, with results presented in \Cref{table:performance_train_data}. The four datasets provide diverse text annotations: LVIS uses simple class names, RefCOCOg contains single-object referring expressions, gRefCOCO includes expressions that may refer to multiple objects, and LISA++ features texts requiring reasoning. Our experiments demonstrate that these datasets consistently improve model performance.

\textbf{Visual QA Ability.}  We also compare {\ours}'s VQA \citep{masry2022chartqa,mathew2021docvqa,liu2024ocrbench,yue2024mmmu,xai2024grok15v} ability with Qwen2VL \citep{wang2024qwen2vl} and the baseline model Qwen2.5VL\citep{bai2025qwen25vl}. As shown in \Cref{table:vqa_results}, {\ours} achieves a slight performance gain even though we do not train on VQA data.

\textbf{Sampling Number.}
\Cref{fig:sampling_num} presents performance comparison with different sampling number. Models are trained using all 7k training samples. We observe an initial performance gain followed by a notable decline with larger sampling number, suggesting that excessive sampling may induce overfitting to the training distribution and consequently degrade generalization capability.

\textbf{Reasoning.}
\Cref{fig:reasoning} compares the performance of models with and without reasoning.
We eliminate reasoning process by removing thinking instruction and thinking reward. Models are trained using all 7k training samples. Our results show that both approaches outperform the baseline. And the reasoning-enhanced model demonstrates significant gain on intricate reasoning segmentation data.

\subsection{Human Evaluation on Reasoning Process}

\begin{wrapfigure}{t}{0.4\textwidth}
  \vspace{-1em}
  \centering
  \setlength{\tabcolsep}{4pt}{
  \begin{minipage}{\linewidth}
    \centering  
    \captionof{table}{Reasoning Process Analysis by IoU Range. IC: Image Consistency; AC: Answer Consistency.}
    \footnotesize
    \label{tab:reasoning_trace}
    \begin{tabular}{
        r
        S[table-format=3.0]
        S[table-format=3.1]
        S[table-format=3.1]
    }
    \toprule
    \textbf{IoU\_Range} & {\textbf{Num}} & {\textbf{IC (\%)}} & {\textbf{AC (\%)}} \\
    \midrule
    0--0.25    & 26  & 76.9   & 46.2 \\
    0.25--0.50 & 29  & 100.0  & 93.1 \\
    0.50--0.75 & 35  & 100.0  & 91.4 \\
    0.75--1.00 & 110  & 100.0 & 100.0 \\
    \midrule
    \textbf{ALL} & 200 & 97.0  & 90.5 \\
    \bottomrule
    \end{tabular}
  \end{minipage}}
  \vspace{-1em}
\end{wrapfigure}
We employ three experts to conduct human evaluations on ReasonSeg-val to assess both answer consistency and image consistency of the reasoning trace. Image consistency measures whether the reasoning trace accurately describes the visual content in the image. Answer consistency evaluates whether the objects and information mentioned in the reasoning trace are consistent with the final predicted output. These evaluations help assess the faithfulness and reliability of the model’s internal reasoning.

We evaluate the reasoning traces at different levels based on IoU (between prediction and ground-truth) values and results are shown in \Cref{tab:reasoning_trace}. We find that the overall image consistency accuracy and answer consistency accuracy reach 97.0\% and 90.5\%, respectively. The majority of problematic reasoning traces are concentrated in cases where the IoU is below 0.25. These results demonstrate that the reasoning trace of {\ours} is accurate and well-grounded, even though the model was trained without human-annotated reasoning data.

\begin{wrapfigure}{t}{0.5\textwidth}
    \centering
    \vspace{-1.5em}
    \includegraphics[width=1.\linewidth]{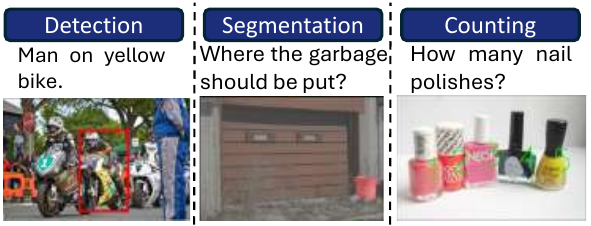}
    \caption{Qualitative results on different tasks. Reasoning process and more results are provided in \Cref{appdx:qualitative}.}
    \vspace{-3em}
    \label{fig:qualitative_simple}
\end{wrapfigure}

\subsection{Qualitative Results}
We visualize some results on \Cref{fig:qualitative_simple}. Notably, {\ours} addresses several visual perception tasks within a shared model. More results and the reasoning process are provided in \Cref{appdx:qualitative}.

\section{Conclusion}

We present {\ours}, a unified vision-language framework for reasoning visual perception tasks.
By introducing novel multi-object cognitive learning strategies and curated reward functions, {\ours} demonstrates strong capabilities in analyzing visual inputs, generating structured reasoning processes and delivering task-specific outputs. 
Experiments across ten diverse tasks, spanning detection, segmentation and counting, validates the robustness and versatility of our approach.
Notably, {\ours} achieves significant improvements over baseline, with relative performance gains of 29.1\% on COCO (detection), 22.1\% on ReasonSeg (segmentation), and 13.2\% on CountBench (counting). 
Human evaluation further reveals the reasoning traces of {\ours} are well-grounded and faithful.

\clearpage
\section*{Acknowledgement}
This work was supported in part by the Research Grants Council under the Areas of Excellence scheme grant AoE/E-601/22-R.

\bibliography{iclr2026_conference}

@misc{xai2024grok15v,
  author = {xAI},
  title = {Grok-1.5 Vision},
  howpublished = {\url{https://x.ai/news/grok-1.5v}},
  year = {2024},
}

@inproceedings{yue2024mmmu,
  title={Mmmu: A massive multi-discipline multimodal understanding and reasoning benchmark for expert agi},
  author={Yue, Xiang and Ni, Yuansheng and Zhang, Kai and Zheng, Tianyu and Liu, Ruoqi and Zhang, Ge and Stevens, Samuel and Jiang, Dongfu and Ren, Weiming and Sun, Yuxuan and others},
  booktitle={Proceedings of the IEEE/CVF Conference on Computer Vision and Pattern Recognition},
  pages={9556--9567},
  year={2024}
}

@article{liu2024ocrbench,
  title={OCRBench: on the hidden mystery of OCR in large multimodal models},
  author={Liu, Yuliang and Li, Zhang and Huang, Mingxin and Yang, Biao and Yu, Wenwen and Li, Chunyuan and Yin, Xu-Cheng and Liu, Cheng-Lin and Jin, Lianwen and Bai, Xiang},
  journal={Science China Information Sciences},
  volume={67},
  number={12},
  pages={220102},
  year={2024},
  publisher={Springer}
}

@inproceedings{kamath2021mdetr,
  title={Mdetr-modulated detection for end-to-end multi-modal understanding},
  author={Kamath, Aishwarya and Singh, Mannat and LeCun, Yann and Synnaeve, Gabriel and Misra, Ishan and Carion, Nicolas},
  booktitle={Proceedings of the IEEE/CVF international conference on computer vision},
  pages={1780--1790},
  year={2021}
}

@article{li2021reftr,
  title={Referring transformer: A one-step approach to multi-task visual grounding},
  author={Li, Muchen and Sigal, Leonid},
  journal={Advances in neural information processing systems},
  volume={34},
  pages={19652--19664},
  year={2021}
}

@inproceedings{deng2021transvg,
  title={Transvg: End-to-end visual grounding with transformers},
  author={Deng, Jiajun and Yang, Zhengyuan and Chen, Tianlang and Zhou, Wengang and Li, Houqiang},
  booktitle={Proceedings of the IEEE/CVF International Conference on Computer Vision},
  pages={1769--1779},
  year={2021}
}

@inproceedings{da2023vgtr,
  title={Vision grid transformer for document layout analysis},
  author={Da, Cheng and Luo, Chuwei and Zheng, Qi and Yao, Cong},
  booktitle={Proceedings of the IEEE/CVF international conference on computer vision},
  pages={19462--19472},
  year={2023}
}

@inproceedings{li2022glip,
  title={Grounded language-image pre-training},
  author={Li, Liunian Harold and Zhang, Pengchuan and Zhang, Haotian and Yang, Jianwei and Li, Chunyuan and Zhong, Yiwu and Wang, Lijuan and Yuan, Lu and Zhang, Lei and Hwang, Jenq-Neng and others},
  booktitle={Proceedings of the IEEE/CVF conference on computer vision and pattern recognition},
  pages={10965--10975},
  year={2022}
}

@inproceedings{liu2023dqdetr,
  title={DQ-DETR: Dual query detection transformer for phrase extraction and grounding},
  author={Liu, Shilong and Huang, Shijia and Li, Feng and Zhang, Hao and Liang, Yaoyuan and Su, Hang and Zhu, Jun and Zhang, Lei},
  booktitle={Proceedings of the AAAI Conference on Artificial Intelligence},
  pages={1728--1736},
  year={2023}
}

@inproceedings{cheng2024yoloworld,
  title={Yolo-world: Real-time open-vocabulary object detection},
  author={Cheng, Tianheng and Song, Lin and Ge, Yixiao and Liu, Wenyu and Wang, Xinggang and Shan, Ying},
  booktitle={Proceedings of the IEEE/CVF Conference on Computer Vision and Pattern Recognition},
  pages={16901--16911},
  year={2024}
}

@article{chen2023shikra,
  title={Shikra: Unleashing multimodal llm's referential dialogue magic},
  author={Chen, Keqin and Zhang, Zhao and Zeng, Weili and Zhang, Richong and Zhu, Feng and Zhao, Rui},
  journal={arXiv preprint arXiv:2306.15195},
  year={2023}
}

@inproceedings{yang2022lavt,
  title={Lavt: Language-aware vision transformer for referring image segmentation},
  author={Yang, Zhao and Wang, Jiaqi and Tang, Yansong and Chen, Kai and Zhao, Hengshuang and Torr, Philip HS},
  booktitle={Proceedings of the IEEE/CVF conference on computer vision and pattern recognition},
  pages={18155--18165},
  year={2022}
}

@inproceedings{liu2023gres,
  title={Gres: Generalized referring expression segmentation},
  author={Liu, Chang and Ding, Henghui and Jiang, Xudong},
  booktitle={Proceedings of the IEEE/CVF conference on computer vision and pattern recognition},
  pages={23592--23601},
  year={2023}
}

@inproceedings{ren2024pixellm,
  title={Pixellm: Pixel reasoning with large multimodal model},
  author={Ren, Zhongwei and Huang, Zhicheng and Wei, Yunchao and Zhao, Yao and Fu, Dongmei and Feng, Jiashi and Jin, Xiaojie},
  booktitle={Proceedings of the IEEE/CVF Conference on Computer Vision and Pattern Recognition},
  pages={26374--26383},
  year={2024}
}

@inproceedings{rasheed2024glamm,
  title={Glamm: Pixel grounding large multimodal model},
  author={Rasheed, Hanoona and Maaz, Muhammad and Shaji, Sahal and Shaker, Abdelrahman and Khan, Salman and Cholakkal, Hisham and Anwer, Rao M and Xing, Eric and Yang, Ming-Hsuan and Khan, Fahad S},
  booktitle={Proceedings of the IEEE/CVF Conference on Computer Vision and Pattern Recognition},
  pages={13009--13018},
  year={2024}
}

@article{li2024llavaov,
  title={Llava-onevision: Easy visual task transfer},
  author={Li, Bo and Zhang, Yuanhan and Guo, Dong and Zhang, Renrui and Li, Feng and Zhang, Hao and Zhang, Kaichen and Zhang, Peiyuan and Li, Yanwei and Liu, Ziwei and others},
  journal={arXiv preprint arXiv:2408.03326},
  year={2024}
}

@misc{r1v,
  author    = {R1-V Team},
  title     = {{R1-V}},
  howpublished = {\url{https://github.com/Deep-Agent/R1-V?tab=readme-ov-file}},
  year      = {2025}
}

@misc{openai2023chatgpt,
  author = {OpenAI},
  title = {ChatGPT},
  year = {2023},
  howpublished = {\url{https://chat.openai.com}},
}

@misc{zheng2025easyr1,
  title        = {EasyR1: An Efficient, Scalable, Multi-Modality RL Training Framework},
  author    = {Zheng, Yaowei and Lu, Junting and Wang, Shenzhi and Feng, Zhangchi and Kuang, Dongdong and Xiong, Yuwen},
  howpublished = {\url{https://github.com/hiyouga/EasyR1}},
  year         = {2025}
}

@article{schulman2017ppo,
  title={Proximal policy optimization algorithms},
  author={Schulman, John and Wolski, Filip and Dhariwal, Prafulla and Radford, Alec and Klimov, Oleg},
  journal={arXiv preprint arXiv:1707.06347},
  year={2017}
}

@article{rafailov2023dpo,
  title={Direct preference optimization: Your language model is secretly a reward model},
  author={Rafailov, Rafael and Sharma, Archit and Mitchell, Eric and Manning, Christopher D and Ermon, Stefano and Finn, Chelsea},
  journal={Advances in Neural Information Processing Systems},
  volume={36},
  pages={53728--53741},
  year={2023}
}

@article{ouyang2022instructgpt,
  title={Training language models to follow instructions with human feedback},
  author={Ouyang, Long and Wu, Jeffrey and Jiang, Xu and Almeida, Diogo and Wainwright, Carroll and Mishkin, Pamela and Zhang, Chong and Agarwal, Sandhini and Slama, Katarina and Ray, Alex and others},
  journal={Advances in neural information processing systems},
  volume={35},
  pages={27730--27744},
  year={2022}
}

@article{zhong2024lyra,
  title={Lyra: An Efficient and Speech-Centric Framework for Omni-Cognition},
  author={Zhong, Zhisheng and Wang, Chengyao and Liu, Yuqi and Yang, Senqiao and Tang, Longxiang and Zhang, Yuechen and Li, Jingyao and Qu, Tianyuan and Li, Yanwei and Chen, Yukang and others},
  journal={arXiv preprint arXiv:2412.09501},
  year={2024}
}

@article{li2024mgm,
  title={Mini-gemini: Mining the potential of multi-modality vision language models},
  author={Li, Yanwei and Zhang, Yuechen and Wang, Chengyao and Zhong, Zhisheng and Chen, Yixin and Chu, Ruihang and Liu, Shaoteng and Jia, Jiaya},
  journal={arXiv preprint arXiv:2403.18814},
  year={2024}
}

@misc{llama32,
  title={Llama 3.2: Revolutionizing edge AI and vision with open, customizable models},  
  author={Meta},  
  year={2024},  
  howpublished={\url{https://ai.meta.com/blog/llama-3-2-connect-2024-vision-edge-mobile-devices/}},  
}

@article{liu2023llava,
  title={Visual instruction tuning},
  author={Liu, Haotian and Li, Chunyuan and Wu, Qingyang and Lee, Yong Jae},
  journal={Advances in neural information processing systems},
  volume={36},
  pages={34892--34916},
  year={2023}
}

@article{guo2025deepseekr1,
  title={Deepseek-r1: Incentivizing reasoning capability in llms via reinforcement learning},
  author={Guo, Daya and Yang, Dejian and Zhang, Haowei and Song, Junxiao and Zhang, Ruoyu and Xu, Runxin and Zhu, Qihao and Ma, Shirong and Wang, Peiyi and Bi, Xiao and others},
  journal={arXiv preprint arXiv:2501.12948},
  year={2025}
}

@article{liu2025visualrft,
  title={Visual-rft: Visual reinforcement fine-tuning},
  author={Liu, Ziyu and Sun, Zeyi and Zang, Yuhang and Dong, Xiaoyi and Cao, Yuhang and Duan, Haodong and Lin, Dahua and Wang, Jiaqi},
  journal={arXiv preprint arXiv:2503.01785},
  year={2025}
}

@inproceedings{peng2024kosmos,
  title={Grounding multimodal large language models to the world},
  author={Peng, Zhiliang and Wang, Wenhui and Dong, Li and Hao, Yaru and Huang, Shaohan and Ma, Shuming and Ye, Qixiang and Wei, Furu},
  booktitle={The Twelfth International Conference on Learning Representations},
  year={2024}
}

@misc{google2025gemini,
  title={Gemini Pro 2.5},  
  author={Google},  
  year={2025},  
  howpublished={\url{https://deepmind.google/technologies/gemini/}},  
}

@misc{openai2025gpt41,
  title={Introducing GPT-4.1 in the API},  
  author={OpenAI},  
  year={2025},  
  howpublished={\url{https://openai.com/index/gpt-4-1/}},  
}

@inproceedings{minderer2022owlvit,
  title={Simple open-vocabulary object detection},
  author={Minderer, Matthias and Gritsenko, Alexey and Stone, Austin and Neumann, Maxim and Weissenborn, Dirk and Dosovitskiy, Alexey and Mahendran, Aravindh and Arnab, Anurag and Dehghani, Mostafa and Shen, Zhuoran and others},
  booktitle={European conference on computer vision},
  pages={728--755},
  year={2022},
  organization={Springer}
}

@article{ren2024dinox,
  title={Dino-x: A unified vision model for open-world object detection and understanding},
  author={Ren, Tianhe and Chen, Yihao and Jiang, Qing and Zeng, Zhaoyang and Xiong, Yuda and Liu, Wenlong and Ma, Zhengyu and Shen, Junyi and Gao, Yuan and Jiang, Xiaoke and others},
  journal={arXiv preprint arXiv:2411.14347},
  year={2024}
}

@article{wang2024qwen2vl,
  title={Qwen2-vl: Enhancing vision-language model's perception of the world at any resolution},
  author={Wang, Peng and Bai, Shuai and Tan, Sinan and Wang, Shijie and Fan, Zhihao and Bai, Jinze and Chen, Keqin and Liu, Xuejing and Wang, Jialin and Ge, Wenbin and others},
  journal={arXiv preprint arXiv:2409.12191},
  year={2024}
}

@misc{cocoapi,
  author = {COCO Team},
  title = {Microsoft {COCO}: Common Objects in Context},
  howpublished = {\url{https://github.com/cocodataset/cocoapi}},
  year = {2014}
}

@article{ravi2024sam2,
  title={Sam 2: Segment anything in images and videos},
  author={Ravi, Nikhila and Gabeur, Valentin and Hu, Yuan-Ting and Hu, Ronghang and Ryali, Chaitanya and Ma, Tengyu and Khedr, Haitham and R{\"a}dle, Roman and Rolland, Chloe and Gustafson, Laura and others},
  journal={arXiv preprint arXiv:2408.00714},
  year={2024}
}

@article{yang2024qwen25,
  title={Qwen2. 5 technical report},
  author={Yang, An and Yang, Baosong and Zhang, Beichen and Hui, Binyuan and Zheng, Bo and Yu, Bowen and Li, Chengyuan and Liu, Dayiheng and Huang, Fei and Wei, Haoran and others},
  journal={arXiv preprint arXiv:2412.15115},
  year={2024}
}

@inproceedings{liu2023grefcoco,
  title={Gres: Generalized referring expression segmentation},
  author={Liu, Chang and Ding, Henghui and Jiang, Xudong},
  booktitle={Proceedings of the IEEE/CVF conference on computer vision and pattern recognition},
  pages={23592--23601},
  year={2023}
}

@article{shao2024deepseekmath,
  title={Deepseekmath: Pushing the limits of mathematical reasoning in open language models},
  author={Shao, Zhihong and Wang, Peiyi and Zhu, Qihao and Xu, Runxin and Song, Junxiao and Bi, Xiao and Zhang, Haowei and Zhang, Mingchuan and Li, YK and Wu, Y and others},
  journal={arXiv preprint arXiv:2402.03300},
  year={2024}
}

@article{liu2025segzero,
  title={Seg-zero: Reasoning-chain guided segmentation via cognitive reinforcement},
  author={Liu, Yuqi and Peng, Bohao and Zhong, Zhisheng and Yue, Zihao and Lu, Fanbin and Yu, Bei and Jia, Jiaya},
  journal={arXiv preprint arXiv:2503.06520},
  year={2025}
}

@article{masry2022chartqa,
  title={Chartqa: A benchmark for question answering about charts with visual and logical reasoning},
  author={Masry, Ahmed and Long, Do Xuan and Tan, Jia Qing and Joty, Shafiq and Hoque, Enamul},
  journal={arXiv preprint arXiv:2203.10244},
  year={2022}
}

@inproceedings{mathew2021docvqa,
  title={Docvqa: A dataset for vqa on document images},
  author={Mathew, Minesh and Karatzas, Dimosthenis and Jawahar, CV},
  booktitle={Proceedings of the IEEE/CVF winter conference on applications of computer vision},
  pages={2200--2209},
  year={2021}
}

@article{deitke2024molmo,
  title={Molmo and pixmo: Open weights and open data for state-of-the-art multimodal models},
  author={Deitke, Matt and Clark, Christopher and Lee, Sangho and Tripathi, Rohun and Yang, Yue and Park, Jae Sung and Salehi, Mohammadreza and Muennighoff, Niklas and Lo, Kyle and Soldaini, Luca and others},
  journal={arXiv preprint arXiv:2409.17146},
  year={2024}
}

@inproceedings{paiss2023countbench,
  title={Teaching clip to count to ten},
  author={Paiss, Roni and Ephrat, Ariel and Tov, Omer and Zada, Shiran and Mosseri, Inbar and Irani, Michal and Dekel, Tali},
  booktitle={Proceedings of the IEEE/CVF International Conference on Computer Vision},
  pages={3170--3180},
  year={2023}
}

@inproceedings{lai2024lisa,
  title={Lisa: Reasoning segmentation via large language model},
  author={Lai, Xin and Tian, Zhuotao and Chen, Yukang and Li, Yanwei and Yuan, Yuhui and Liu, Shu and Jia, Jiaya},
  booktitle={Proceedings of the IEEE/CVF Conference on Computer Vision and Pattern Recognition},
  pages={9579--9589},
  year={2024}
}

@article{yang2023lisa++,
  title={LISA++: An Improved Baseline for Reasoning Segmentation with Large Language Model},
  author={Yang, Senqiao and Qu, Tianyuan and Lai, Xin and Tian, Zhuotao and Peng, Bohao and Liu, Shu and Jia, Jiaya},
  journal={arXiv preprint arXiv:2312.17240},
  year={2023}
}

@inproceedings{yu2016refcoco,
  title={Modeling context in referring expressions},
  author={Yu, Licheng and Poirson, Patrick and Yang, Shan and Berg, Alexander C and Berg, Tamara L},
  booktitle={Computer Vision--ECCV 2016: 14th European Conference, Amsterdam, The Netherlands, October 11-14, 2016, Proceedings, Part II 14},
  pages={69--85},
  year={2016},
  organization={Springer}
}

@inproceedings{kazemzadeh2014referitgame,
  title={Referitgame: Referring to objects in photographs of natural scenes},
  author={Kazemzadeh, Sahar and Ordonez, Vicente and Matten, Mark and Berg, Tamara},
  booktitle={Proceedings of the 2014 conference on empirical methods in natural language processing (EMNLP)},
  pages={787--798},
  year={2014}
}

@inproceedings{gupta2019lvis,
  title={Lvis: A dataset for large vocabulary instance segmentation},
  author={Gupta, Agrim and Dollar, Piotr and Girshick, Ross},
  booktitle={Proceedings of the IEEE/CVF conference on computer vision and pattern recognition},
  pages={5356--5364},
  year={2019}
}

@inproceedings{lin2014mscoco,
  title={Microsoft coco: Common objects in context},
  author={Lin, Tsung-Yi and Maire, Michael and Belongie, Serge and Hays, James and Perona, Pietro and Ramanan, Deva and Doll{\'a}r, Piotr and Zitnick, C Lawrence},
  booktitle={Computer vision--ECCV 2014: 13th European conference, zurich, Switzerland, September 6-12, 2014, proceedings, part v 13},
  pages={740--755},
  year={2014},
  organization={Springer}
}

@article{bai2025qwen25vl,
  title={Qwen2. 5-vl technical report},
  author={Bai, Shuai and Chen, Keqin and Liu, Xuejing and Wang, Jialin and Ge, Wenbin and Song, Sibo and Dang, Kai and Wang, Peng and Wang, Shijie and Tang, Jun and others},
  journal={arXiv preprint arXiv:2502.13923},
  year={2025}
}

@inproceedings{liu2024grounding,
  title={Grounding dino: Marrying dino with grounded pre-training for open-set object detection},
  author={Liu, Shilong and Zeng, Zhaoyang and Ren, Tianhe and Li, Feng and Zhang, Hao and Yang, Jie and Jiang, Qing and Li, Chunyuan and Yang, Jianwei and Su, Hang and others},
  booktitle={European Conference on Computer Vision},
  pages={38--55},
  year={2024},
  organization={Springer}
}

@article{yu2025dapo,
  title={Dapo: An open-source llm reinforcement learning system at scale},
  author={Yu, Qiying and Zhang, Zheng and Zhu, Ruofei and Yuan, Yufeng and Zuo, Xiaochen and Yue, Yu and Dai, Weinan and Fan, Tiantian and Liu, Gaohong and Liu, Lingjun and others},
  journal={arXiv preprint arXiv:2503.14476},
  year={2025}
}

@misc{internvl2,
  title        = {Introduction of InternVL2 Series},
  author       = {OpenGVLab},
  url          = {https://internvl.readthedocs.io/en/latest/internvl2.0/introduction.html},
  year         = {2024},
  urldate      = {2024}
}

@article{sheng2024verl,
  title   = {HybridFlow: A Flexible and Efficient RLHF Framework},
  author  = {Guangming Sheng and Chi Zhang and Zilingfeng Ye and Xibin Wu and Wang Zhang and Ru Zhang and Yanghua Peng and Haibin Lin and Chuan Wu},
  year    = {2024},
  journal = {arXiv preprint arXiv: 2409.19256}
}
\bibliographystyle{iclr2026_conference}

\clearpage
\appendix
\section{The Use of Large Language Models (LLMs)}
LLMs are used only to polish writing in this paper.

\begin{figure}[t]
  \centering
   \includegraphics[width=1.0\linewidth]{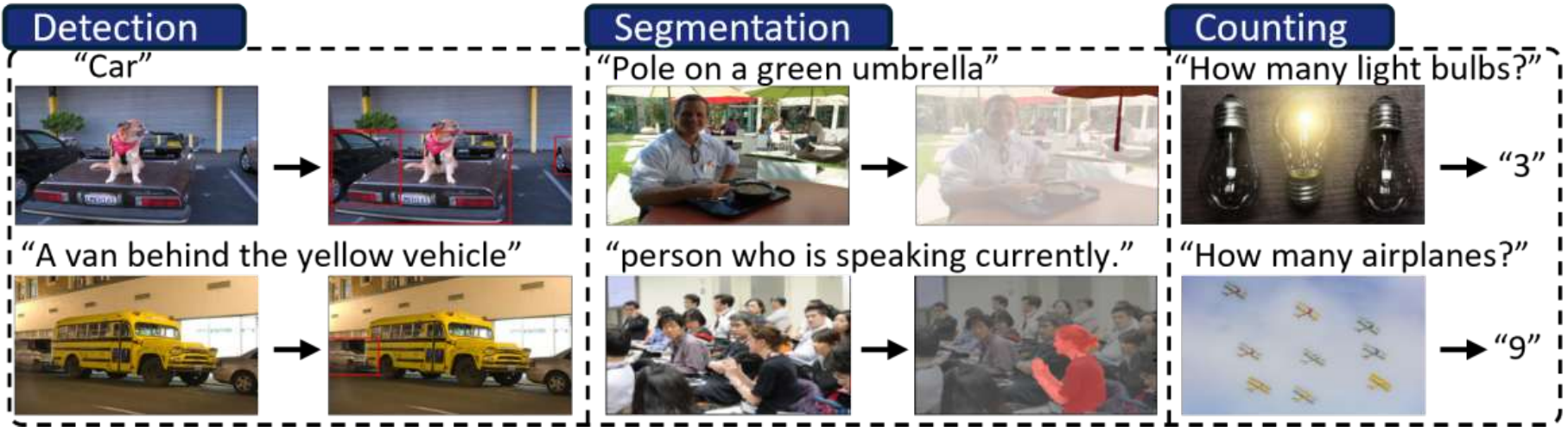}
   \caption{Examples from evaluation benchmarks. Zoom in for better viewing.}
   \label{fig:uvp_visual}
\end{figure}

\section{Details of Evaluation Benchmarks}
\label{appdx:detail_benchmark}

\begin{wrapfigure}{t}{0.5\textwidth}
  \vspace{-1em}
  \centering
  \setlength{\tabcolsep}{9pt}{
  \begin{minipage}[t]{\linewidth}
    \centering  
    \footnotesize
    \captionof{table}{Statistics of evaluation benchmarks. We report the number of instances for detection and segmentation tasks. The reported numbers combine validation and test splits where applicable.}
    \label{tab:uvp_bench}
    \begin{tabular}{llr}
      \toprule
      \textbf{Type} & \textbf{Data} & \textbf{\# of samples} \\
      \midrule
      \multirow{4}{*}{Det} & COCO      & 36,781 \\
                          & RefCOCO   & 5,786 \\ 
                           & RefCOCO+  & 5,060 \\ 
                           & RefCOCOg  & 7,596 \\ 
      \midrule
      \multirow{4}{*}{Seg} & RefCOCO   & 1,975 \\ 
                           & RefCOCO+  & 1,975 \\ 
                           & RefCOCOg  & 5,023 \\ 
                            & ReasonSeg & 979 \\ 
      \midrule
      \multirow{2}{*}{Count} & Pixmo-Count & 1,064 \\ 
                            & CountBench & 504 \\ 
      \midrule
      \textbf{SUM} &  & 66,023 \\
      \bottomrule
    \end{tabular}
  \end{minipage}}
  \vspace{-1em}
\end{wrapfigure}
We use ten benchmarks to evaluate model performance across general vision perception tasks. Our evaluation includes three fundamental task types: detection, segmentation and counting. Specifially, we employ COCO \citep{lin2014mscoco} and RefCOCO(+/g) \citep{yu2016refcoco} for detection evaluation; RefCOCO(+/g) and ReasonSeg \citep{lai2024lisa} for segmentation evaluation; PixMo-Count \citep{deitke2024molmo} and CountBench \citep{paiss2023countbench} for counting.

\textbf{Annotation Preparation.} To ensure consistency across all evaluation tasks, we standardize the evaluation data by converting all samples into a unified multi-modal conversation format and removing potential information leakage. This preprocessing involves: converting numeric class labels to textual descriptions in COCO \citep{lin2014mscoco}; removing explicit numerical references from text descriptions in CountBench \citep{paiss2023countbench}; applying consistent formatting across all datasets to maintain evaluation fairness.

\textbf{Evaluation Metrics.} For object detection on COCO, we adopt the standard AP metric computed using the COCO API \citep{cocoapi}. For referring object grounding on RefCOCO(+/g), we use bbox AP, which measures detection accuracy at an IoU threshold of 0.5. For object segmentation on RefCOCO(+/g) and ReasonSeg, we use gIoU, computed as the mean IoU across all segmentation masks. 
For counting tasks, we use count accuracy as evaluation metric. 

\textbf{Statistics and Visualization.} We show the statistic data  in  \Cref{tab:uvp_bench}. For detection and segmentation tasks, we report the number of valid instances. For counting tasks, we provide the total number of test samples. Our evaluation comprises a total of 66,023 test samples, covering three fundamental visual perception task types and 10 specific tasks. We visualize some examples in \Cref{fig:uvp_visual}.

\section{More Training Details}
The training is conducted on a single node with 8 GPUs and the entire training process takes 6 hours. The peak GPU memory usage is approximately 80 GB, though this can be adjusted through hyperparameters such as memory\_utilization in VeRL \citep{sheng2024verl}. The reward converges at around 100 steps, and the best checkpoint is typically obtained at around 200 steps.

\section{Experission Level Evaluation on RefCOCO(+/g)}
Our primary evaluation, detailed in \Cref{appdx:detail_benchmark}, reports instance-level performance. However, since the RefCOCO(+/g) benchmarks provide multiple expressions per image, we additionally present expression-level results in \Cref{table:performance_detection_expressionlevel}.

\setlength{\tabcolsep}{8.5pt}{
\begin{table}[h]
\centering
\caption{Performance comparison on expression-level RefCOCO(+/g) tasks. Results with * are cited from the Qwen2.5-VL report but are not reproducible in our environment.}
\footnotesize
\begin{tabular}{l|cc|cc|cc|c}
\toprule
\multirow{4}{*}{\textbf{Method}} & \multicolumn{6}{c}{\textbf{Detection}} \vline & \multirow{4}{*}{\textbf{SUM}} \\
\cmidrule{2-7} 
     & \multicolumn{2}{c}{RefCOCO} \vline & \multicolumn{2}{c}{RefCOCO+} \vline & \multicolumn{2}{c}{RefCOCOg} \vline  \\
\cmidrule{2-7} 
      & val & testA &  val & testA & val & test \\
\midrule
\textcolor{gray}{Qwen2.5-VL-7B*}  & \textcolor{gray}{90.0} & \textcolor{gray}{92.5} & \textcolor{gray}{84.2} & \textcolor{gray}{89.1} & \textcolor{gray}{87.2} & \textcolor{gray}{87.2} & \textcolor{gray}{530.2} \\
Qwen2.5-VL-7B  & 89.0 & 92.0 & 83.2 & 88.3 & 86.4 & 86.5 & 525.4 \\
\cellcolor[HTML]{efefef}{{\ours}-7B}  & \cellcolor[HTML]{efefef}{89.1} & \cellcolor[HTML]{efefef}{91.0} & \cellcolor[HTML]{efefef}{85.0} & \cellcolor[HTML]{efefef}{87.6} & \cellcolor[HTML]{efefef}{87.6} & \cellcolor[HTML]{efefef}{88.5} & \cellcolor[HTML]{efefef}{\textbf{528.8}} \\
\bottomrule
\end{tabular}
\label{table:performance_detection_expressionlevel}
\end{table}}




\section{Task Router}

In order to identify users' instruction automatically during inference, we also train a TaskRouter.
The TaskRouter $\mathcal{F}_{\text{router}}$ is a pure language model that processes textual instructions. For any given instruction $\mathbf{T}$, TaskRouter performs a semantic analysis and outputs a task classification $\mathbf{C}$ into one of four predefined fundamental task categories. This mapping can be formally expressed as:

\begin{equation}
    \mathbf{C} = \mathcal{F}_{\text{router}}(\mathbf{T}).
\end{equation}

\begin{wrapfigure}{t}{0.3\textwidth}
  \vspace{-3pt}
  \centering
  \setlength{\tabcolsep}{5pt}{
  \begin{minipage}{\linewidth}
    \centering  
    \captionof{table}{Comparison on the task classification.}
    \footnotesize
    \label{tab:task_router}
    \begin{tabular}{l|c}
      \toprule
      \textbf{Model} & \textbf{Accuracy} \\
      \midrule
             Qwen2.5-1.5B      & 46.3 \\
             \cellcolor[HTML]{efefef}{TaskRouter-1.5B}  & \cellcolor[HTML]{efefef}{\textbf{99.1}} \\
      \bottomrule
    \end{tabular}
  \end{minipage}}
\end{wrapfigure}

We train TaskRouter using the GRPO algorithm \citep{shao2024deepseekmath}, providing reward signals exclusively upon correct task classification.
We evaluate the effective of the TaskRouter and results are shown on \Cref{tab:task_router}. The task classification dataset is constructed from diverse visual perception datasets and AI-generated samples. For datasets that include textual instructions (\textit{e.g.}, RefCOCOg), we retain their original instructions and corresponding fundamental task categories. 
Additionally, for each fundamental task type, we employ ChatGPT \citep{openai2023chatgpt} to generate instructions and target categories. 
The final dataset comprises 20,000 training samples and 4,000 test samples.
Although the state-of-the-art Qwen2.5 \citep{yang2024qwen25} demonstrates strong performance in instruction following and zero-shot task classification, its accuracy drops below 50\% in our complex scenario. In contrast, our task router module, trained using reinforcement learning, achieves significantly better performance.

\section{User Prompt Template}
To guide the policy model toward generating desired outputs during exploration, we employ the user prompt template presented in \Cref{tab:object_matching_template}. This prompt template is inspired by DeepSeek-R1-Zero \citep{guo2025deepseekr1} and Seg-Zero \citep{liu2025segzero}.

\definecolor{darkblue}{RGB}{28,46,121}
\begin{table}[h]
\centering
\tcbset{
  colback=gray!5!white,    
  colframe=darkblue,           
  width=1.\linewidth,    
  boxrule=1pt,             
  arc=4mm,                 
  left=5pt,                
  right=5pt,               
  top=5pt,                 
  bottom=5pt,              
}
\caption{User Prompt. ``\textit{\{Question\}}'' is replaced by user questions during training and inference.}
\footnotesize 
\begin{tcolorbox}[title=User Prompt]

"Please find ``\textit{\{Question\}}'' with bboxs and points." \\
"Compare the difference between object(s) and find the most closely matched object(s)." \\
"Output the thinking process in <think> </think> and final answer in <answer> </answer> tags." \\
"Output the bbox(es) and point(s) inside the interested object(s) in JSON format."
\begin{verbatim}
i.e. <think> thinking process here </think>
<answer>[{"bbox_2d": [10,100,200,210], "point_2d" [30,110]}, 
{"bbox_2d": [225,296,706,786], "point_2d": [302,410]}]</answer>
\end{verbatim}

\normalsize
\label{tab:object_matching_template}
\end{tcolorbox}
\end{table}

\begin{figure}[t]
    \centering
    \includegraphics[width=1.0\linewidth]{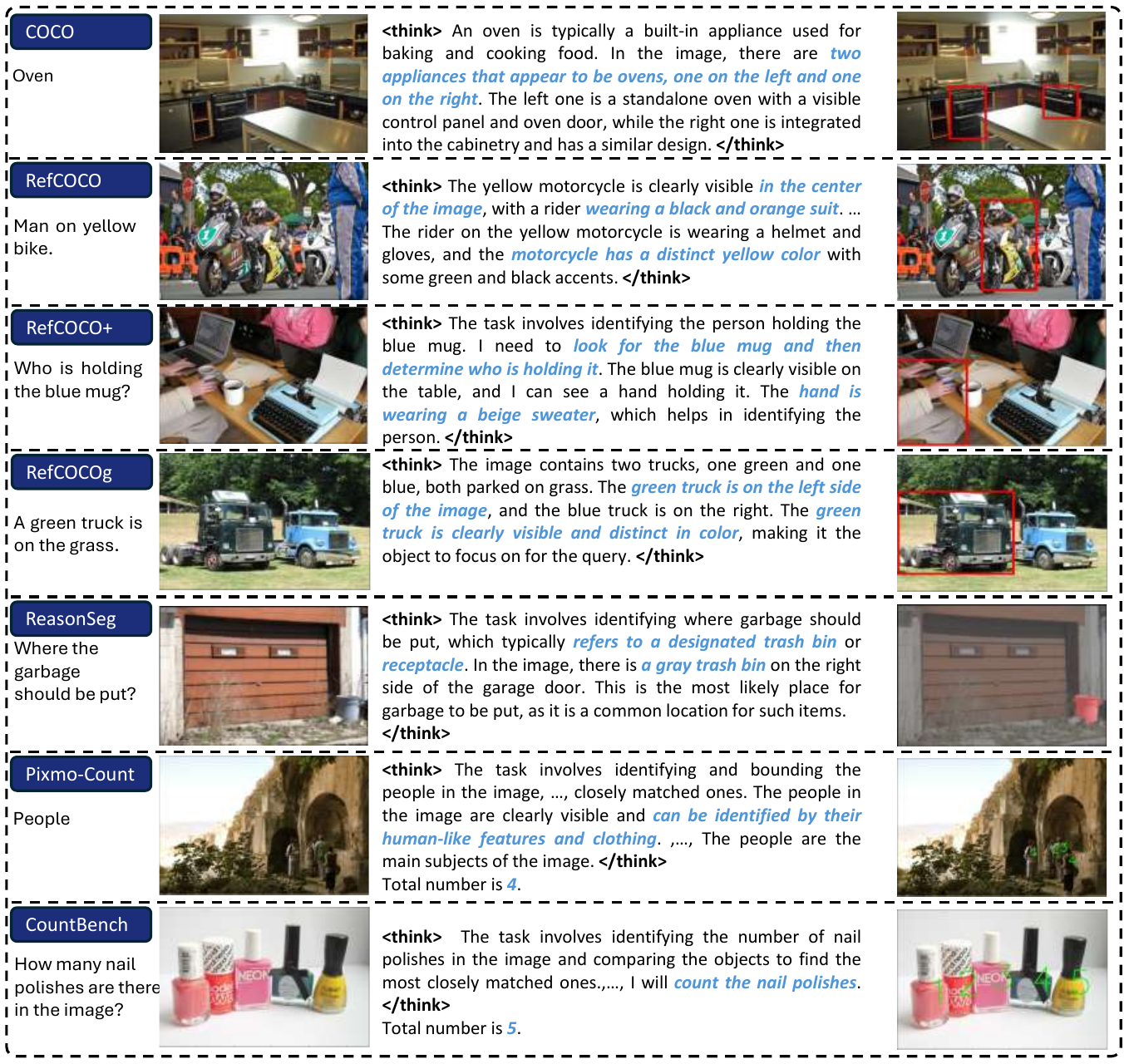}
    \caption{Qualitative results on different tasks. Zoom in for better visualization.}
    \label{fig:qualitative}
\end{figure}

\section{Qualitative Results}
\label{appdx:qualitative}
We visualize results on \Cref{fig:qualitative}. 
Our model generates comprehensive reasoning processes for all tasks while producing expected outputs. We find that {\ours} can effectively distinguish between similar objects, as shown in the visual grounding and referring expression segmentation. {\ours} also accurately localize multiple targets, as shown in object detection and counting. We also observe that the length of the reasoning process adapts dynamically: more intricate image-query pairs elicit detailed rationales, while simpler inputs result in concise explanations.

\begin{figure}[t]
    \centering
    \includegraphics[width=1.0\linewidth]{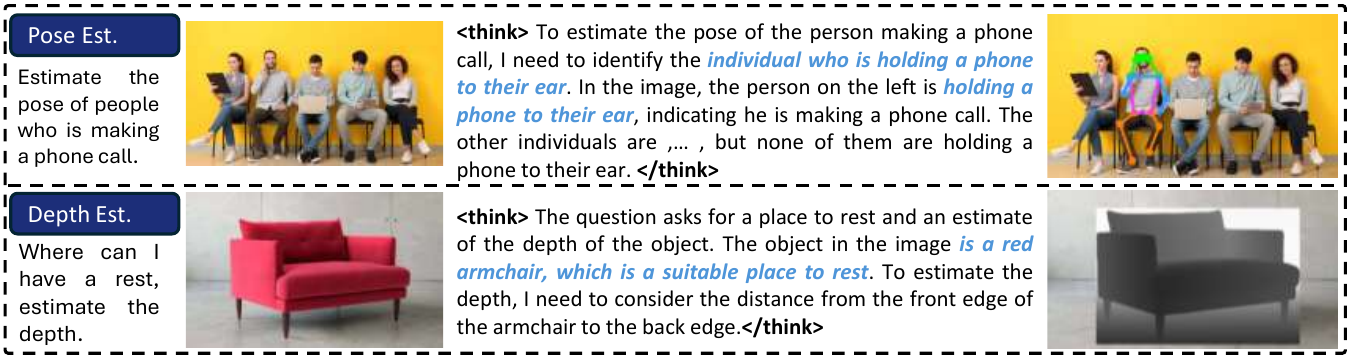}
    \caption{Extending {\ours} to more visual perception tasks.}
    \label{fig:more_task}
\end{figure}

\section{Extension}
\label{appdx:extension}
\textbf{More Applications.} Our {\ours} can be easily extended to other visual perception tasks that need reasoning. We just need to add a light-weight module for different output format. The intermediate output $\{\mathbf{B}_i\}_{i=1}^{N}$ and $\{\mathbf{P}_i\}_{i=1}^{N}$ serve as bridge to connect other modules. \Cref{fig:more_task} shows our extension to referring expression pose estimation and referring expression depth estimation.

\textbf{Hybrid Mode.} We can also employ a hybrid mode, that is directly using traditional visual models (e.g. Yolo-World \citep{cheng2024yoloworld}) for simple categorical instruction (i.e. bird) and {\ours} for complex instructions (i.e. `Where can I have a rest?').

\section{Concepts Clarification}
We formally define the key terms used in this work. As illustrated in Table~\ref{tab:concepts}, our hierarchical task formulation adopts the COCO dataset \citep{lin2014mscoco} as a representative example:

\begin{table}[ht]
\centering
\caption{Key Terminology and Definitions.}
\tablestyle{2.9pt}{1.1}
\resizebox{1.\linewidth}{!}{
\begin{tabular}{y{120}|y{260}}
\toprule
\textbf{Concept} & \textbf{Definition} \\
\toprule
Fundamental Task Types & Reformulated task categories (\textit{e.g.}, detection) \\
\midrule
Task Type & Task category (\textit{e.g.}, object detection) \\
\midrule
Task & Concrete benchmark (\textit{e.g.}, COCO object detection) \\
\bottomrule
\end{tabular}}
\label{tab:concepts}
\end{table}

\section{Details of Task Reformulation}
\label{appdx:task_reformulation}

Within our framework, we categorize task types as illustrated in \Cref{tab:counting_vqa_type} and \Cref{tab:detection_segmentation_type}. These tables highlight our grouping of task types based on their similarities. It is important to note that although this taxonomy covers a broad range of task types, the current implementation of {\ours} is evaluated on only 10 representative tasks, with comprehensive evaluation of all task types reserved for future research.

\begin{table}[ht]
\centering
\caption{Fundamental Task Types: Counting and Visual Question Answering.}
\footnotesize
\tablestyle{2.9pt}{1.1}
\resizebox{1.\linewidth}{!}{
\begin{tabular}{y{190}|y{190}}
\toprule
\textbf{Counting} & \textbf{VQA} \\
\midrule
Object Counting & Visual Question Answering (VQA) \\
Crowd Counting & Classification \\
Density Estimation & Image Captioning \\
Pedestrian Detection & Question Answering \\
Crowd Estimation in Dense Scenes & Visual Reasoning \\
Traffic Counting in Surveillance & Visual Question Answering \\
 & Relational Reasoning \\
\bottomrule
\end{tabular}}
\label{tab:counting_vqa_type}
\end{table}

\begin{table}[ht]
\centering
\caption{Fundamental task types: Detection and Segmentation.}
\footnotesize
\tablestyle{2.9pt}{1.1}
\resizebox{1.\linewidth}{!}{
\begin{tabular}{y{190}|y{190}}
\toprule
\textbf{Detection} & \textbf{Segmentation} \\
\midrule
Visual Grounding & Semantic Segmentation \\
Object Detection & Instance Segmentation \\
2D Object Detection & Lane Detection \\
Small Object Detection & 2D Semantic Segmentation \\
Defect Detection & Medical Image Segmentation \\
Face Detection & Human Part Segmentation \\
License Plate Detection & Action Segmentation \\
Anomaly Detection & Video Object Segmentation \\
Human Detection & Referring Expression Segmentation \\
Surgical Tool Detection & Saliency Detection \\
Dense Object Detection & Salient Object Detection \\
Open World Object Detection & Semantic Segmentation of Remote Sensing Imagery \\
Zero-Shot Object Detection & Crack Segmentation \\
Animal Action Recognition & Action Unit Detection \\
Robotic Grasping & RGB Salient Object Detection \\
Object Localization & Boundary Detection \\
Hand Detection & Crack Segmentation for Infrastructure \\
Visual Relationship Detection & Surgical Tool Segmentation \\
Open Vocabulary Object Detection &  \\
Oriented Object Detection &  \\
Object Detection in Indoor Scenes &  \\
Object Detection in Aerial Images &  \\
Person Search &  \\
Object Recognition &  \\
\bottomrule
\end{tabular}}
\label{tab:detection_segmentation_type}
\end{table}

\end{document}